\documentclass{article}

\usepackage{PRIMEarxiv}

\usepackage[utf8]{inputenc} 
\usepackage[T1]{fontenc}    
\usepackage{hyperref}       
\usepackage{url}            
\usepackage{booktabs}       
\usepackage{amsfonts}       
\usepackage{nicefrac}       
\usepackage{microtype}      
\usepackage{lipsum}
\usepackage{natbib} 
\usepackage{fancyhdr}       
\usepackage{graphicx}       
\graphicspath{{media/}}     

\usepackage{amsmath}
\usepackage{amssymb}
\usepackage{amsbsy}
\usepackage{amsthm}
\usepackage{algorithm} 
\usepackage{algpseudocode}
\usepackage{booktabs}
\usepackage{siunitx}
\usepackage{natbib}

\title{Graph-Enhanced Deep Reinforcement Learning for Multi-Objective Unrelated Parallel Machine Scheduling
}

\author{
  Bulent Soykan \\
  Institute for Simulation and Training \\
  University of Central Florida \\
  Orlando, FL, USA\\
  \texttt{Bulent.Soykan@ucf.edu} \\
   \And
  Sean Mondesire, Ghaith Rabadi, Grace Bochenek \\
  School of Modeling, Simulation, and Training \\
  University of Central Florida \\
  Orlando, FL, USA\\
  \texttt{\{Sean.Mondesire, Ghaith.Rabadi, Grace.Bochenek\}@ucf.edu} \\
}

\begin{document}
\maketitle

\begin{abstract}
The Unrelated Parallel Machine Scheduling Problem (UPMSP) with release dates, setups, and eligibility constraints presents a significant multi-objective challenge. Traditional methods struggle to balance minimizing Total Weighted Tardiness (TWT) and Total Setup Time (TST). This paper proposes a Deep Reinforcement Learning framework using Proximal Policy Optimization (PPO) and a Graph Neural Network (GNN). The GNN effectively represents the complex state of jobs, machines, and setups, allowing the PPO agent to learn a direct scheduling policy. Guided by a multi-objective reward function, the agent simultaneously minimizes TWT and TST. Experimental results on benchmark instances demonstrate that our PPO-GNN agent significantly outperforms a standard dispatching rule and a metaheuristic, achieving a superior trade-off between both objectives. This provides a robust and scalable solution for complex manufacturing scheduling.
 Source code and data are available at: \href{https://github.com/bulentsoykan/GNN-DRL4UPMSP}{github.com/bulentsoykan/GNN-DRL4UPMSP}.
\end{abstract}


\section{INTRODUCTION}
\label{sec:intro}

Advanced manufacturing environments operate under intense pressure to improve efficiency, reduce operational costs, and maintain responsiveness to dynamic market demands \cite{davis2012smart}. The implementation of efficient production scheduling is crucial for achieving these goals. It involves finding the best way to allocate constrained resources, such as machines and personnel, to competing tasks or jobs over a specific timeframe. Optimal scheduling decisions directly impact key performance indicators (KPIs), including production throughput, resource utilization, adherence to delivery deadlines, and overall system productivity. Consequently, developing adaptive scheduling methodologies is important for maintaining competitiveness in today's globalized industrial landscape \cite{mourtzis2022advances}.

A fundamental class of scheduling problems encountered in various industrial settings is the Parallel Machine Scheduling Problem (PMSP). In this context, a set of jobs must be processed on a set of available machines. This paper focuses on a particularly challenging variant known as the Unrelated Parallel Machine Scheduling Problem (UPMSP). In the UPMSP, the processing time required for a specific job depends not only on the job itself but also distinctly on the machine assigned to process it. This scenario accurately reflects real-world situations where machines, even if performing similar functions, may differ in age, capability, or specialization. Such problems are prevalent in diverse sectors, including semiconductor manufacturing \cite{bitar2016memetic}, textile production \cite{li2021hybrid}, where heterogeneity among processing units is common.

Real-world UPMSP applications are further complicated by several interacting constraints addressed in this research. These include job-specific release dates, machine eligibility restrictions, and setup times that depend on both the sequence of jobs and the specific machine. Handling these complexities simultaneously significantly increases the difficulty of finding optimal or near-optimal schedules. In addition, effective scheduling must often balance conflicting objectives, such as minimizing total weighted tardiness (TWT) to meet deadlines and minimizing total setup time (TST) for operational efficiency. The inherent conflict—where prioritizing one objective negatively impacts the other—makes finding high-performing solutions challenging. Given the $\mathsf{NP}$-hard nature of this multi-objective UPMSP \cite{fleszar2018algorithms}, traditional approaches face limitations in effectively managing this trade-off.

In this paper, we propose a novel method based on Deep Reinforcement Learning (DRL) in order to address the limitations of existing methods and tackle the complexities of the multi-objective UPMSP. DRL agents learn decision-making policies through interaction with an environment, making them well-suited for sequential decision problems like scheduling. We specifically use Graph Neural Networks (GNNs) to handle the intricate structure and variable nature of the scheduling state. GNNs can effectively process graph-based representations, capturing the relationships between jobs, machines, eligibility constraints, and setup dependencies. Our framework utilizes a heterogeneous GNN representation designed to encode the rich information inherent in the UPMSP state, including distinct node types for jobs, machines, and potential setup configurations, connected by meaningful edge types. This GNN acts as a powerful feature extractor, feeding state information to a DRL agent trained using the Proximal Policy Optimization (PPO) algorithm. PPO is chosen for its stability and proven effectiveness in complex control tasks, enabling the agent to learn a policy that directly selects which job to assign to which machine at each decision step.

The main contributions of this research are threefold. First, we propose and evaluate the application of the PPO algorithm for learning a \textit{direct scheduling policy} for the complex UPMSP variant, moving beyond heuristic selection or simpler scheduling scenarios often addressed by DRL. Second, we design and implement a \textit{heterogeneous GNN state representation} tailored to the UPMSP, explicitly modeling jobs, machines, setup states, and their complex interrelations to facilitate more effective learning. Third, we develop and analyze an \textit{explicit multi-objective reward function} within the PPO framework, specifically designed to guide the agent in learning policies that effectively balance the conflicting objectives.

The remainder of this paper is organized as follows. Section \ref{sec:related_work} provides a review of relevant literature on UPMSP, DRL in scheduling, and GNN applications. Section \ref{sec:problem_formulation} formally defines the target UPMSP and its objectives. Section \ref{sec:methodology} details the proposed PPO-GNN framework, including the MDP formulation, GNN architecture, and reward function design. Section \ref{sec:experiments} describes the experimental setup, instance generation, baseline methods, and evaluation metrics. Section \ref{sec:results} presents and discusses the computational results and comparative analysis. Finally, Section \ref{sec:conclusion} concludes the paper and outlines directions for future research.

\section{Related Work}
\label{sec:related_work}

\subsection{Parallel Machine Scheduling Problem (PMSP)}
\label{subsec:related_pmsp}
The PMSP encompasses a broad category of scheduling challenges where multiple jobs need to be processed using a set of available machines operating in parallel. Research in PMSP branches into several classifications based on machine characteristics: identical machines (where processing times are job-dependent only), uniform machines (where machines have different speeds but maintain relative processing time proportions), and unrelated machines (where processing time depends arbitrarily on both the job and the machine) \cite{mokotoff2001parallel}. This work focuses on the \textit{UPMSP}, which poses significant combinatorial challenges due to the machine-dependent nature of processing times.

The complexity of UPMSP grows significantly when incorporating realistic industrial constraints. A substantial body of literature addresses UPMSP variants that include sequence-dependent setup times, where the time to prepare a machine depends on the previous job processed \cite{vallada2011genetic}, machine-dependent setup times inherent to the unrelated nature \cite{avalos2015efficient}, distinct release dates for jobs \cite{li2022novel}, and machine eligibility restrictions limiting which machines can process certain jobs \cite{afzalirad2018design}. Common objectives studied include minimizing the makespan ($C_{\max}$), total completion time ($\sum C_j$), total tardiness ($\sum T_j$), and TWT ($\sum w_j T_j$) \cite{mokotoff2001parallel}. Recent research also increasingly considers setup time minimization as a critical objective due to its impact on resource utilization and efficiency. This paper addresses the UPMSP with release dates, sequence- and machine-dependent setup times, and machine eligibility, aiming to simultaneously minimize TWT and TST.

Due to the inherent $\mathsf{NP}$-hard computational complexity of most UPMSP variants, \textit{heuristics and dispatching rules} are widely employed, particularly in dynamic environments requiring rapid decision-making \cite{rabadi2006heuristics}. These methods prioritize jobs waiting for processing based on specific rules, such as Shortest Processing Time (SPT), Earliest Due Date (EDD), or Critical Ratio (CR). For more complex scenarios involving setup times and tardiness objectives, composite rules like the Apparent Tardiness Cost (ATC) rule and its derivatives, such as Apparent Tardiness Cost with Setups (ATCS), have been developed \cite{lee1997heuristic}. Variations like Apparent Tardiness Cost with Setups and Ready Times (ATCSR) explicitly incorporate release dates and setup times \cite{lin2014unrelated}. While computationally efficient, these rules typically make greedy, myopic decisions based on local information. Their performance can degrade significantly when faced with the intricate interactions present in UPMSP with sequence-dependent setups and multiple objectives, often leading to suboptimal solutions \cite{rabadi2006heuristics}.

In order to overcome the limitations of simple heuristics, various metaheuristic and exact approaches have been proposed. \textit{Metaheuristics} such as Genetic Algorithms (GA) \cite{vallada2011genetic}, Tabu Search (TS) \cite{lee2013tabu}, Simulated Annealing (SA) \cite{kim2002unrelated}, or Ant Colony Optimization (ACO) \cite{afzalirad2016design} explore the solution space more broadly and can yield higher-quality solutions compared to simple heuristics. However, they typically require considerable problem-specific parameter tuning, can be computationally intensive, provide no guarantee of optimality, and their iterative nature may limit their applicability in highly dynamic or real-time decision-making scenarios \cite{moser2022exact}. \textit{Exact methods} such as Mixed-Integer Programming (MIP) \cite{saracc2022mix}, Constraint Programming (CP) \cite{gedik2018constraint}, or specialized Dynamic Programming (DP) algorithms can guarantee optimal solutions \cite{pfund2004survey}. Unfortunately, due to the combinatorial complexity of the problem, these methods suffer from the "curse of dimensionality". As a result, they are typically only practical to compute for very small problem sizes, making them unsuitable for application at realistic industrial scales \cite{pei2020new}. These limitations highlight the need for alternative approaches that can effectively handle the scale and complexity of the target problem while learning to navigate the intricate trade-offs between multiple objectives.

\subsection{Deep Reinforcement Learning in Scheduling}
\label{subsec:related_drl_scheduling}
Recently, DRL has become known as an effective technique for tackling complex sequential decision-making problems, including combinatorial optimization problems such as scheduling \cite{mazyavkina2021reinforcement}. Unlike traditional optimization methods that often search through a vast solution space, DRL aims to learn a \textit{policy}—a mapping from the current system state to an optimal or near-optimal action—through trial-and-error interactions with an environment (often a simulation model) \cite{soykanICMLA2023}. The integration of deep neural networks allows DRL agents to handle high-dimensional state spaces and learn complex patterns, making them suitable for capturing the dynamics of intricate scheduling environments \cite{cunha2020deep}.

DRL has been successfully applied to various classical scheduling problems. For the Job Shop Scheduling Problem (JSSP), significant research has focused on learning dispatching policies \cite{soykanWintersim2024}. Early work often involved using DRL to select from a predefined set of dispatching rules \cite{zhang2020learning}. More recent approaches utilize advanced state representations, such as disjunctive graphs processed by GNNs, enabling agents to learn end-to-end policies that directly select the next operation to schedule \cite{park2021learning}. Similar advancements have been made for the Flow Shop Scheduling Problem (FSPP), including permutation flow shops \cite{cho2022minimize} and flexible flow shops \cite{kwon2021matrix}, often employing sequence-to-sequence models or GNNs to handle job dependencies and routing decisions. Objectives commonly tackled include makespan minimization and tardiness reduction. These studies demonstrate the potential of DRL to learn high-quality, adaptive scheduling strategies directly from simulated experience.

PPO \cite{schulman2017proximal} is a state-of-the-art policy gradient DRL algorithm known for its stability, reliability, and strong practical performance across a wide range of challenging tasks, including continuous control and combinatorial optimization problems. Compared to simpler policy gradient methods like REINFORCE or actor-critic methods like A2C/A3C, PPO incorporates a clipping mechanism in the objective function or an adaptive Kullback-Leibler (KL) penalty to constrain policy updates, preventing excessively large changes that can destabilize training. This makes it attractive for scheduling problems where the state-action space is large and finding a good policy requires careful exploration. Its balance between sample efficiency, ease of implementation, and robust performance motivates its selection as the core learning algorithm in our proposed framework for the multi-objective UPMSP.

\subsection{Graph Neural Networks in Combinatorial Optimization}
\label{subsec:related_gnn_co}

GNNs have emerged as a highly effective deep learning architecture for combinatorial optimization problems, many of which naturally lend themselves to graph representations where nodes signify entities and edges capture relationships or dependencies \cite{huang2019review}. GNNs utilize message-passing mechanisms to iteratively aggregate information from node neighborhoods, learning powerful embeddings that encode both node features and the graph's topology, making them adept at handling variable-sized inputs common in combinatorial optimization \cite{chung2025neural}. Their success, often in conjunction with DRL, is evident across various combinatorial optimization domains, including routing problems, where GNN encoders generate representations for sequence-constructing decoders \cite{kool2018attention}, and other fundamental problems like SAT and MaxCut \cite{huang2019review}. Given that scheduling problems can frequently be modeled as graphs (e.g., disjunctive graphs for JSSP), GNNs are increasingly central to modern DRL-based scheduling approaches for JSSP \cite{zhang2020learning} and FSPP \cite{kwon2021matrix}. For PMSP, Cho et al.  specifically utilized a GNN to represent the state (including machine-job pair nodes and machine nodes) for solving the UPMSP with setups, release dates, and eligibility, focusing on minimizing TWT using the REINFORCE algorithm \cite{cho2024reinforcement}.  Norman et al. also employed a GNN in a scheduling context, but used it primarily as a feature extractor to inform a DRL agent that tuned parameters within a predefined dispatching rule, rather than performing direct scheduling \cite{norman2024}. These works underscore the GNN's ability to effectively encode the complex relational information present in scheduling environments.

While previous research has demonstrated the potential of DRL and GNNs for various scheduling problems, several gaps remain, particularly concerning the complex, multi-objective UPMSP variant addressed here. Existing DRL approaches for UPMSP often focus on single objectives like TWT \cite{cho2024reinforcement}, select from predefined heuristics rather than learning direct policies \cite{nam2024}, or tune parameters of existing rules \cite{norman2024}. Also, while GNNs have been used, the design of graph representations tailored to capture the information needed to effectively balance TWT minimization against TST minimization in UPMSP requires further investigation.

This paper endeavors to bridge these gaps by introducing a novel framework that uniquely combines several elements: \textit{(i)} we employ the robust \textit{PPO} algorithm to learn a \textit{direct scheduling policy} for the UPMSP, offering potential stability advantages over simpler policy gradient methods used in some prior work, \textit{(ii)} we tackle the \textit{multi-objective nature} of the problem explicitly, designing a reward function and learning process aimed at simultaneously minimizing both TWT and TST, \textit{(iii)} we introduce an \textit{enhanced heterogeneous GNN structure} specifically designed for this multi-objective UPMSP variant, incorporating distinct nodes for jobs, machines, and potential setups, along with carefully defined edge types to capture eligibility, processing times, and setup relationships (including transitions).
By integrating these components, our work seeks to advance the state-of-the-art in applying DRL to solve large-scale, complex, multi-objective scheduling problems encountered in real-world manufacturing.


\section{Problem Formulation}
\label{sec:problem_formulation}
This section formally defines the UPMSP variant addressed in this paper, including its parameters, decision variables, constraints, and objectives.

\paragraph{Formal Definition}
\label{subsec:formal_definition}
We consider a set of independent jobs $J = \{J_1, J_2, \dots, J_n\}$ that need to be processed on a set of unrelated parallel machines $M = \{M_1, M_2, \dots, M_m\}$. The problem parameters are defined as follows:
\begin{itemize}
    \item $p_{jk}$: The processing time of job $J_j \in J$ when processed on machine $M_k \in M$. The machines are unrelated, meaning $p_{jk}$ can vary arbitrarily for different machines $k$ and jobs $j$.
    \item $r_j$: The release date (or ready time) of job $J_j \in J$, representing the earliest time at which processing of job $J_j$ can begin.
    \item $d_j$: The due date of job $J_j \in J$, representing the target completion time for the job.
    \item $w_j$: The weight or importance of job $J_j \in J$, used in calculating the weighted tardiness objective.
    \item $s_{ijk}$: The sequence-dependent and machine-dependent setup time required on machine $M_k \in M$ to switch from processing job $J_i \in J \cup \{J_0\}$ to processing job $J_j \in J$. Here, $J_0$ represents the initial idle state of the machine, and $s_{0jk}$ denotes the initial setup time if job $J_j$ is the first job processed on machine $M_k$.
    \item $M_j \subseteq M$: The set of eligible machines on which job $J_j \in J$ can be processed. If $M_k \notin M_j$, job $J_j$ cannot be assigned to machine $M_k$.
\end{itemize}

\paragraph{Decision Variables}
\label{subsec:decision_variables}
The primary decisions involved in solving this UPMSP instance are (These decisions implicitly determine the start time ($S_{jk}$), completion time ($C_{jk}$), and ultimately the overall schedule performance based on the defined objectives. Let $C_j$ denote the completion time of job $J_j$ on the machine it is assigned to.):
\begin{itemize}
    \item \textit{Job Assignment:} Determining which eligible machine $M_k \in M_j$ will process each job $J_j \in J$.
    \item \textit{Job Sequencing:} Determining the order (sequence) in which the jobs assigned to each specific machine $M_k \in M$ will be processed.
\end{itemize}

\paragraph{Constraints}
\label{subsec:constraints}
The scheduling decisions must adhere to the following constraints:
\begin{itemize}
    \item \textit{Release Date:} Processing for job $J_j$ on its assigned machine $M_k$ cannot start before its release date: $S_{jk} \ge r_j$.
    \item \textit{Machine Eligibility:} Each job $J_j$ must be assigned to a machine $M_k$ from its eligible set: $M_k \in M_j$.
    \item \textit{Non-preemption:} Once the processing of a job starts on a machine, it cannot be interrupted until it is completed.
    \item \textit{Machine Capacity:} Each machine $M_k$ can process at most one job at any given time. If job $J_j$ immediately follows job $J_i$ in the sequence on machine $M_k$, then the start time of $J_j$ must be greater than or equal to the completion time of $J_i$ plus the required setup time: $S_{jk} \ge C_{ik} + s_{ijk}$.
\end{itemize}

\paragraph{Objective Functions}
\label{subsec:objective_functions}
The goal is to find a feasible schedule that simultaneously optimizes two conflicting objectives:

\begin{itemize}
    \item \textit{Minimize Total Weighted Tardiness (TWT)}
\label{subsubsec:obj_twt}
Tardiness for job $J_j$ is defined as $T_j = \max(0, C_j - d_j)$. The objective is to minimize the sum of weighted tardiness over all jobs:
\begin{equation}
    \label{eq:twt}
    \text{Minimize } TWT = \sum_{j=1}^{n} w_j T_j = \sum_{j=1}^{n} w_j \max(0, C_j - d_j)
\end{equation}

 \item \textit{Minimize Total Setup Time (TST)}
\label{subsubsec:obj_setup}
Let $\sigma_k$ be the sequence of jobs assigned to machine $M_k$, represented as $(J_{k(1)}, J_{k(2)}, \dots, J_{k(n_k)})$, where $n_k$ is the number of jobs assigned to $M_k$. The TST is the sum of all setup times incurred across all machines:
\begin{equation}
    \label{eq:setup_time}
    \text{Minimize } TST = \sum_{k=1}^{m} \left( s_{0k(1)k} + \sum_{l=1}^{n_k-1} s_{k(l)k(l+1)k} \right)
\end{equation}
where $s_{0k(1)k}$ is the setup time for the first job $J_{k(1)}$ on machine $M_k$.
\end{itemize}

\section{Methodology: PPO-GNN for Multi-Objective UPMSP}
\label{sec:methodology}
This section details the proposed DRL framework for solving the multi-objective UPMSP. We use PPO combined with a tailored GNN architecture. 

\begin{figure}[h]
    \centering
    \includegraphics[width=0.5\textwidth]{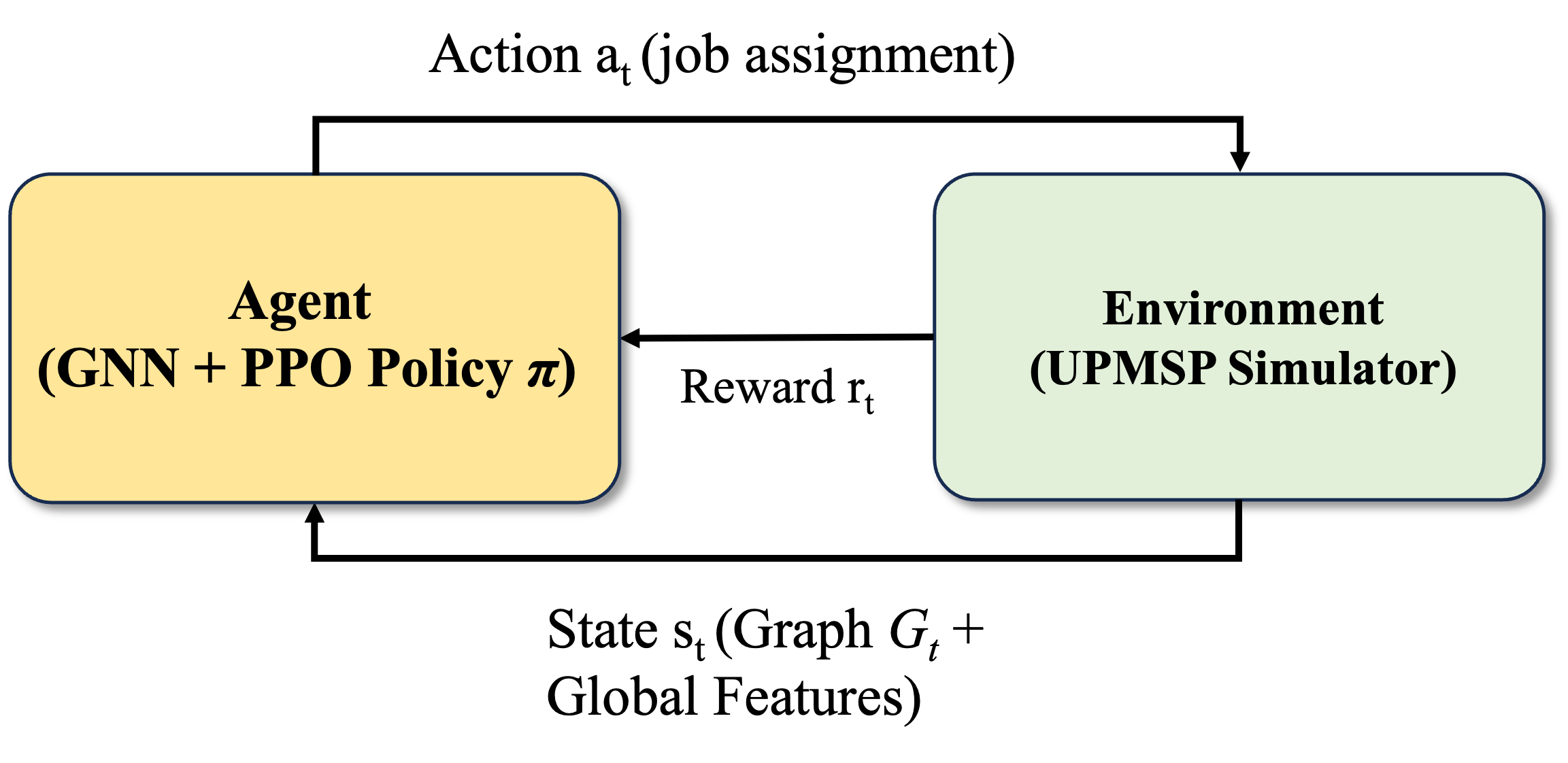}

    \caption{Overview of the DRL framework for UPMSP scheduling.}
    \label{fig:framework}
    \vspace{-3mm}

\end{figure}

The core of our approach is an interaction loop between a DRL agent and a scheduling environment, implemented as a discrete-event simulator. At specific decision points (e.g., when a machine becomes idle), the environment provides the current state $s_t$ to the agent. The agent, using its learned policy $\pi$, selects an action $a_t$ (assigning a job to a machine or waiting). The environment executes this action, advancing the simulation time and determining the next state $s_{t+1}$ and an immediate reward $r_t$. The agent collects these experiences $(s_t, a_t, r_t, s_{t+1})$ to update its policy and value function parameters via the PPO algorithm, aiming to maximize the cumulative discounted reward over time. Figure \ref{fig:framework} illustrates this interaction loop.

\subsection{Markov Decision Process (MDP) Formulation}
\label{subsec:mdp_formulation}
We formulate the UPMSP as a Markov Decision Process (MDP), defined by the tuple $(\mathcal{S}, \mathcal{A}, \mathcal{P}, \mathcal{R}, \gamma)$, where $\mathcal{S}$ is the state space, $\mathcal{A}$ is the action space, $\mathcal{P}$ is the state transition probability function (implicitly defined by the simulator), $\mathcal{R}$ is the reward function, and $\gamma$ is the discount factor.

\paragraph{State Representation ($s_t \in \mathcal{S}$)}
\label{subsubsec:state}
Capturing the complex state of the UPMSP environment effectively is important for the DRL agent. We propose a composite state representation $s_t$ at decision time step $t$, consisting of a global feature vector and a heterogeneous graph:

 \noindent\textbf{Global Feature Vector:} Provides a summary of the overall system status, including features: (i) current number of jobs waiting in the queue (WIP).
            (ii) number of jobs expected to arrive within a near-future time horizon.
            (iii) aggregate performance metrics (e.g., number of currently tardy jobs, average flow time of completed jobs, TST incurred so far).
            (iv) overall machine status (e.g., number of idle machines, number of machines under setup, number of machines undergoing maintenance if applicable).
            (v) current simulation time and time elapsed.

\noindent\textbf{Heterogeneous Graph ($\mathcal{G}_t$):} Represents the detailed relationships between entities in the scheduling environment. It contains different types of nodes and edges:

    \textit{Node Types \& Features:}
                    \textit{(i) Job Nodes:} One node for each job $J_j$ currently waiting or arriving soon. Features include its weight ($w_j$), remaining processing time estimate (average $p_{jk}$ over eligible machines, or specific $p_{jk}$ if assignment is imminent), due date ($d_j$), release date ($r_j$), and potentially features indicating its setup requirements.
                    \textit{(ii) Machine Nodes:} One node for each machine $M_k$. Features include its current status (idle, busy, setup, maintenance), time until available, identifier of the last job processed ($J_i$), current setup state (e.g., identifier corresponding to $J_i$), and time elapsed in the current state/setup.
                    \textit{(iii) Setup Nodes:} Nodes representing distinct and relevant setup configurations (e.g., characterized by job family, required tooling, or energy level). Features identify the setup type. These nodes allow summarizing information about jobs requiring a specific setup, even if no machine currently has it.
          
\textit{Edge Types \& Features:} Edges represent relationships and potential actions. Features encode relevant costs and properties:
                    \textit{(i) Job-Machine Edges:} Connect job node $J_j$ to machine node $M_k$ if $M_k \in M_j$. Features include processing time $p_{jk}$, estimated setup time $s_{i'jk}$ (where $J_{i'}$ is the last job on $M_k$), eligibility flag (binary).
                    \textit{(ii) Machine-Setup Edges:} Connect machine node $M_k$ to the setup node representing its current setup state. Features indicate time spent in this setup.
                    \textit{(iii) Job-Setup Edges:} Connect job node $J_j$ to the setup node representing the setup required before processing $J_j$. Features indicate job priority ($w_j$).
                    \textit{(iv) Setup-Machine Edges:} Connect setup node $S$ to machine node $M_k$. Features represent the estimated time/cost to transition machine $M_k$ from its current setup to setup $S$, indicating 'transition ease' (e.g., small setup time if only a minor change is needed).

\paragraph{Action Space ($a_t \in \mathcal{A}$)}
\label{subsubsec:action}
The action space consists of potential scheduling decisions. The primary action is the selection of a feasible job-machine pair $(J_j, M_k)$ to be scheduled next. The agent's policy $\pi(a_t|s_t)$ provides a probability distribution over the feasible actions.
\begin{itemize}
    \item A pair $(J_j, M_k)$ is feasible if job $J_j$ is available (i.e., $r_j$ is met or will be met by the time the machine is ready), machine $M_k$ is eligible for $J_j$ ($M_k \in M_j$), and $M_k$ is currently idle or will become idle.
    \item The agent's output (logits from the policy network) corresponds to all possible job-machine pairs. An action masking mechanism is applied to ensure only feasible actions have non-zero selection probability. Infeasible actions (e.g., assigning to an ineligible machine, assigning a job that is not released) are masked out (given zero or highly negative log-probability).
\end{itemize}

\paragraph{Reward Function ($r_t = \mathcal{R}(s_t, a_t, s_{t+1})$)}
Designing the reward function $r_t$ is critical for steering the DRL agent to effectively balance the conflicting objectives of minimizing TWT and TST. We explored several structures to achieve this balance, including dense, event-based rewards calculated after each assignment $a_t=(J_j, M_k)$ (e.g., $r_t = - \alpha \cdot \Delta TWT_j - \beta \cdot s_{i'jk}$), hybrid approaches combining immediate penalties for setup time with sparser rewards reflecting overall episode performance on both TWT and TST, and constraint-focused designs incorporating penalties for auxiliary KPI violations. The specific formulation and the relative weighting of components (via factors like $\alpha, \beta$) significantly influence the learned policy's trade-off behavior and require careful empirical tuning to achieve the desired multi-objective performance.

\paragraph{State Transition ($\mathcal{P}(s_{t+1}|s_t, a_t)$)}
\label{subsubsec:transition}
The state transition dynamics are implicitly defined by the discrete-event simulation environment. Given the current state $s_t$ and the agent's chosen action $a_t$, the simulator advances time, updates machine statuses, calculates job completion times (including processing and setup), handles new job arrivals based on their release dates, and determines the resulting next state $s_{t+1}$. The DRL agent interacts with this simulator without needing an explicit model of the transition probabilities.

\section{Experimental Setup}
\label{sec:experiments}
This section details the experimental methodology employed to evaluate the performance of our PPO-GNN framework. We describe the generation of problem instances, the baseline methods selected for comparison, the configuration of the DRL agent's training process, and the metrics used for performance evaluation.

\paragraph{Instance Generation}
We generated a diverse set of UPMSP instances following methodologies adapted from established literature, particularly \cite{cho2024reinforcement}. Problem instances were constructed by systematically varying parameters known to influence scheduling complexity. Specifically, we considered job set sizes $n \in \{20, 50, 100\}$ and machine set sizes $m \in \{5, 10, 15\}$. Processing times $p_{jk}$ for each job-machine pair were independently drawn from a discrete uniform distribution $DU(1, 100)$, guaranteeing machine unrelatedness. An overall average processing time $\bar{p}$ was calculated across all potential job-machine pairings for reference in generating other parameters. Job release dates $r_j$ were sampled from $DU(0, \lambda \cdot \bar{p})$, where the arrival intensity factor was set to $\lambda=0.5$. Due dates $d_j$ were determined using standard tightness ($\tau$) and range ($R$) parameters, relative to the job's release date and its average processing time across eligible machines ($\bar{p}_j$), according to $d_j \sim DU(r_j + \bar{p}_j(1-\tau-R/2), r_j + \bar{p}_j(1-\tau+R/2))$. We utilized combinations of $\tau \in \{0.2, 0.4, 0.6\}$ and $R \in \{0.2, 0.6, 1.0\}$. Job importance weights $w_j$ were sampled from $DU(1, 10)$. Sequence-dependent and machine-dependent setup times $s_{ijk}$ were drawn from $DU(0, \beta \cdot \bar{p})$ using setup ratios $\beta \in \{0.1, 0.25\}$, applied randomly without imposing specific job family structures. Machine eligibility was controlled via a density parameter $\delta \in \{0.75, 1.0\}$; each job $J_j$ was randomly assigned to $\lceil \delta \cdot m \rceil$ machines from $M$, ensuring $J_j$ was eligible for at least one machine. For each distinct combination of parameters ($n, m, \tau, R, \beta, \delta$), 50 unique instances were generated to ensure robust statistical analysis.

\paragraph{Baseline Methods}
In order to assess the performance of our PPO-GNN approach, we compared it against representative algorithms from different scheduling methodology classes:

\textbf{(i) Dispatching Rule}
We implemented a well-regarded composite dispatching rule designed for parallel machine environments with setups and due dates: the \textbf{ATCSR\_Rm} rule \cite{lin2014unrelated}. This adaptation of the ATC rule explicitly accounts for setup times and release dates and serves as a strong heuristic baseline due to its demonstrated performance in related UPMSP contexts.

\textbf{(ii) Metaheuristic Algorithm}
We implemented a \textbf{Genetic Algorithm (GA)} tailored for the multi-objective UPMSP, drawing inspiration from approaches like \cite{vallada2011genetic}. The GA employed a chromosome representation encoding job sequences for each machine and utilized standard genetic operators (selection, crossover, mutation). Its fitness function aimed to minimize a weighted sum of normalized objectives: $\alpha \cdot TWT_{norm} + (1-\alpha) \cdot TST_{norm}$. Normalization was performed by dividing the objective values by those obtained from a fast run of the ATCSR\_Rm heuristic ($TWT_{norm} = TWT / TWT_{ATCSR}$, $TST_{norm} = TST / TST_{ATCSR}$). We used a representative weight $\alpha=0.5$ to balance the objectives. The GA was executed with a fixed computational time limit per instance, set to be comparable to the inference time required by the trained DRL agent during evaluation, ensuring a fair comparison of solution quality within similar time constraints. GA parameters (population size, mutation rate, etc.) were tuned via preliminary experiments.

\paragraph{Training Configuration}
The PPO agent, including the GNN feature extractor and MLP policy/value heads, was implemented using PyTorch \cite{paszke2019pytorch} and the Stable Baselines3 library \cite{raffin2021stable}. The GNN architecture comprised 4 layers of the GATv2 message-passing scheme \cite{brody2021attentive} with a hidden dimension of 128 and ReLU activation functions. The actor and critic networks shared the GNN encoder and subsequently utilized separate MLPs, each having 2 hidden layers with 256 neurons and ReLU activations. Key PPO hyperparameters were refined through preliminary tuning: the learning rate $\eta$ was initialized at $1 \times 10^{-4}$ with linear decay over training, the discount factor $\gamma=0.99$, Generalized Advantage Estimation lambda $\lambda_{GAE}=0.95$, PPO clipping parameter $\epsilon=0.2$, number of optimization epochs per data collection phase set to 10, and a mini-batch size of 64. Training was conducted for $10^6$ total environment steps, accelerated using multiple parallel actors interacting with simulator instances. All training and evaluation experiments were executed on a system equipped with an Intel Core i9 CPU and an NVIDIA RTX 3090 GPU.

\paragraph{Evaluation Metrics}
The performance of the trained PPO-GNN agent and all baseline methods was  evaluated on a dedicated set of test instances, kept separate from those used during training or hyperparameter tuning. The primary evaluation metrics directly reflect the multi-objective nature of the problem:
    \textit{(i) Average TWT:} The mean TWT value, as defined in Eq. \ref{eq:twt}, 
    \textit{(ii) Average TST:} The mean TST, as defined in Eq. \ref{eq:setup_time}. Both are calculated across all instances in the test set for each evaluated method.
As a secondary metric reflecting practical usability, we report the \textit{Average Computational Time} (inference/run time) required by each algorithm to produce a schedule for a single test instance during the evaluation phase. We employ \textit{Pareto Front visualizations}, plotting the (Avg TST, Avg TWT) pairs achieved by the different methods to facilitate the analysis of trade-offs between the primary objectives. Statistical significance of the performance differences between our proposed PPO-GNN method and the baselines was assessed using paired t-tests.

\section{Results and Discussion}
\label{sec:results}

\begin{figure}[ht]
    \centering
    \includegraphics[width=0.9\linewidth]{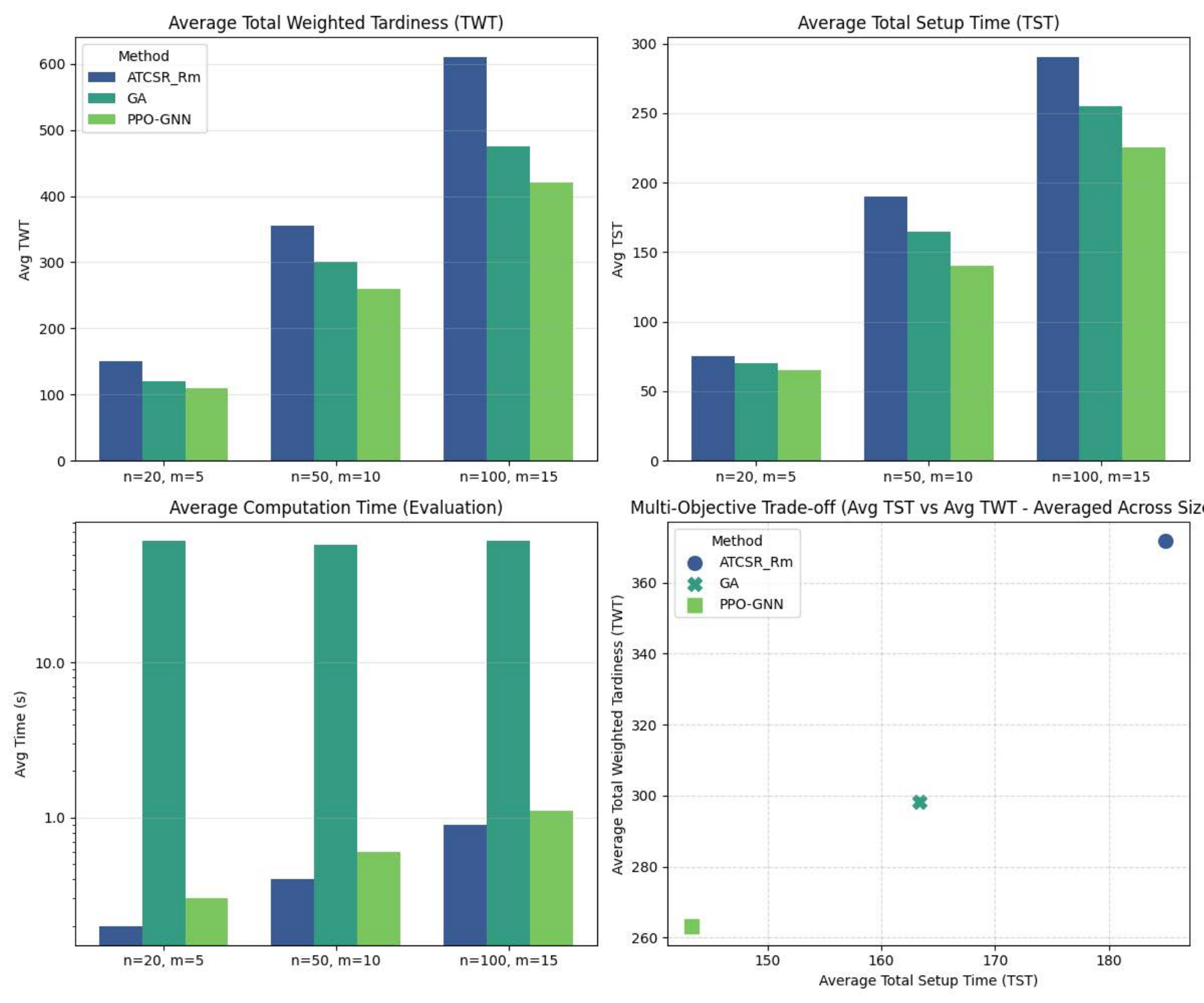}
    \caption{Comparison of scheduling methods across different problem sizes (n jobs, m machines).}
    \label{fig:results}
\end{figure}

This section presents and discusses the computational results obtained by evaluating the proposed PPO-GNN agent against the baseline methods (ATCSR\_Rm and GA) across varying problem sizes. The performance comparison focuses on the primary objectives of minimizing Average TWT (Avg TWT) and Average TST (Avg TST), as well as the evaluation-phase computational efficiency. All results are summarized in Table~\ref{tab:results} and visualized in Figure~\ref{fig:results}.

The results show the superior performance of the proposed PPO-GNN approach in optimizing both primary objectives. As shown in the top row of Figure~\ref{fig:results} and detailed in Table~\ref{tab:results}, the PPO-GNN agent consistently achieved the lowest Avg TWT and the lowest Avg TST across all tested problem sizes (n=20/m=5, n=50/m=10, n=100/m=15). Compared to the ATCSR\_Rm dispatching rule, PPO-GNN yielded substantial reductions in both metrics. While the GA baseline improved upon ATCSR\_Rm, it was consistently outperformed by PPO-GNN. For instance, on the largest instances (n=100, m=15), PPO-GNN achieved an Avg TWT of 420.0 compared to 475.0 for GA and 610.0 for ATCSR\_Rm, while simultaneously achieving an Avg TST of 225.0 compared to 255.0 for GA and 290.0 for ATCSR\_Rm. The bolded values in Table~\ref{tab:results} highlight the best performance achieved for each metric and size, corresponding to the PPO-GNN method. Statistical significance tests (paired t-tests, multiple runs per instance) was performed to confirm these observed differences ($p < 0.01$).

\begin{table}[ht]
\centering
\caption{Performance Comparison of PPO-GNN against Baselines} 
\label{tab:results}
\footnotesize 

\begin{tabular}{@{} ll rrr @{}} 
\toprule
       Size &    Method &           Avg TWT &           Avg TST &  Avg Comp Time (s) \\
\midrule
 n=20, m=5 &  ATCSR\_Rm &             150.0 &              75.0 &              61.29 \\ 
 n=20, m=5 &        GA &             120.0 &              70.0 &               0.10 \\ 
 n=20, m=5 &   PPO-GNN & \textbf{110.0} &   \textbf{65.0} &               0.52 \\ 
 \addlinespace 
 n=50, m=10 &  ATCSR\_Rm &             355.0 &             190.0 &              59.17 \\ 
 n=50, m=10 &        GA &             300.0 &             165.0 &               0.11 \\ 
 n=50, m=10 &   PPO-GNN & \textbf{260.0} & \textbf{140.0} &               0.92 \\ %
 \addlinespace 
n=100, m=15 &  ATCSR\_Rm &             610.0 &             290.0 &              61.29 \\ 
n=100, m=15 &        GA &             475.0 &             255.0 &               0.12 \\ 
n=100, m=15 &   PPO-GNN & \textbf{420.0} & \textbf{225.0} &               1.57 \\ 
\bottomrule
\end{tabular}
\end{table}

The evaluation-phase computational time, presented in Table~\ref{tab:results} and visualized in the bottom-left panel of Figure~\ref{fig:results} (note the logarithmic scale on the Y-axis required to effectively display the range), reveals significant differences in efficiency. As expected, the simple ATCSR\_Rm heuristic is the fastest, executing almost instantaneously. The proposed PPO-GNN agent, while requiring neural network computations, also demonstrates high efficiency with inference times remaining under 1.6 seconds even for the largest problems. In contrast, the GA method requires significantly more time (around 60 seconds) during evaluation to perform its iterative search process. This highlights a crucial practical advantage of the trained DRL agent: its ability to generate high-quality scheduling decisions rapidly upon deployment, making it suitable for dynamic environments where fast responses are often necessary.

The performance of the GA is highly dependent on its parameter settings and the computational time allowed for its search. The GA implemented here was tuned through preliminary experiments to serve as a representative and strong baseline. While further extensive tuning or a significantly longer run time might yield improved results for the GA, this would also underscore a key practical advantage of our DRL approach. The PPO-GNN agent requires a one-time, offline training cost, but its inference during evaluation is exceptionally fast. In contrast, the GA requires a significant computational budget for every new problem instance it solves. Regarding the performance across individual instances, our results in Table 1 and Figure 2 present the average performance, where the PPO-GNN agent is consistently superior. While the performance gap may vary on an instance-by-instance basis, the PPO-GNN method demonstrated strictly better performance on average across all tested problem sizes and configurations. This indicates a robust and generalizable policy rather than a trade-off where one method excels on some instances and not others.

The multi-objective trade-off between minimizing Avg TST and Avg TWT, averaged across all problem sizes, is visualized in the bottom-right panel of Figure~\ref{fig:results}. The axes are oriented such that the ideal performance lies in the bottom-left corner (minimum TST and minimum TWT). The plot shows that the point representing the PPO-GNN agent dominates the points for both GA and ATCSR\_Rm. This means that, on average, PPO-GNN achieves strictly better performance on \textit{both} objectives simultaneously compared to the baselines. It doesn't merely represent a different trade-off point on the Pareto front; rather, it pushes the achieved performance envelope closer to the ideal objective vector. This strong result suggests that the PPO-GNN agent effectively learns complex scheduling strategies that inherently reduce both setup times and tardiness concurrently, likely by making more globally informed decisions facilitated by the GNN's structural awareness.

As expected, both Avg TWT and Avg TST tend to increase for all methods as the problem size (number of jobs 'n' and machines 'm') grows, reflecting the increased complexity and load. However, the \textit{relative} advantage of the PPO-GNN agent appears consistent or even slightly increasing across the tested sizes. This suggests that the learned policy generalizes reasonably well to larger problems within the scope of the training distribution.

The performance of the PPO-GNN framework can be attributed to several factors inherent in its design. The GNN's ability to process the complex graph representation allows it to capture the intricate dependencies between jobs, machines, eligibility, and potential setups more effectively than simpler rule-based logic (ATCSR\_Rm) or population encodings (GA). The PPO algorithm effectively utilizes these rich features to learn a sophisticated policy capable of navigating the complex state space and balancing the conflicting objectives, guided by the carefully designed multi-objective reward signal. Unlike the myopic nature of dispatching rules or the potentially time-consuming search of metaheuristics, the learned DRL policy offers a powerful blend of high-quality decision-making and fast execution speed during deployment.

\section{Conclusion and Future Work}
\label{sec:conclusion}

In this paper, we addressed the multi-objective UPMSP incorporating release dates, sequence- and machine-dependent setup times, and machine eligibility constraints. Recognizing the limitations of traditional heuristics and the computational demands of exact methods and metaheuristics, we proposed a novel DRL framework. Our approach leverages PPO combined with an enhanced heterogeneous GNN designed to effectively capture the intricate state information of the scheduling environment, including job, machine, and setup characteristics. We developed a multi-objective reward function explicitly aimed at guiding the agent to simultaneously minimize TWT and TST. Simulations on a range of generated benchmark instances demonstrated the superiority of the proposed PPO-GNN agent. It outperformed both a dispatching rule (ATCSR\_Rm) and a GA baseline, achieving lower Avg TWT and Avg TST across various problem scales, thereby showcasing its ability to effectively learn high-quality policies for complex, multi-objective scheduling scenarios. The primary contributions of this work include: (i) the successful application of PPO for learning a direct scheduling policy in the challenging multi-objective UPMSP context; (ii) the design of a heterogeneous GNN state representation tailored for capturing UPMSP complexities relevant to balancing tardiness and setup objectives; and (iii) the demonstration that a DRL agent can effectively learn to manage conflicting objectives through an appropriately designed reward structure.

\bibliographystyle{unsrt}  
\bibliography{references}

@ARTICLE{afzalirad2016design,
  title={{Design of High-Performing Hybrid Meta-Heuristics for Unrelated Parallel Machine Scheduling with Machine Eligibility and Precedence Constraints}},
  author={Afzalirad, Mojtaba and Rezaeian, Javad},
  journal={Engineering Optimization},
  volume={48},
  number={4},
  pages={706--726},
  year={2016}
}

@ARTICLE{afzalirad2018design,
  title={{Design of an Efficient Genetic Algorithm for Resource-Constrained Unrelated Parallel Machine Scheduling Problem with Machine Eligibility Restrictions}},
  author={Afzalirad, Mojtaba and Shafipour, Masoud},
  journal={Journal of Intelligent Manufacturing},
  volume={29},
  number={2},
  pages={423--437},
  year={2018}
}

@ARTICLE{avalos2015efficient,
  title={{Efficient Metaheuristic Algorithm and Re-Formulations for the Unrelated Parallel Machine Scheduling Problem with Sequence and Machine-Dependent Setup Times}},
  author={Avalos-Rosales, Oliver and Angel-Bello, Francisco and Alvarez, Ada},
  journal={The International Journal of Advanced Manufacturing Technology},
  volume={76},
  pages={1705--1718},
  year={2015}
}

@ARTICLE{bitar2016memetic,
  title={{A Memetic Algorithm to Solve an Unrelated Parallel Machine Scheduling Problem with Auxiliary Resources in Semiconductor Manufacturing}},
  author={Bitar, Abdoul and Dauz{\`e}re-P{\'e}r{\`e}s, St{\'e}phane and Yugma, Claude and Roussel, Renaud},
  journal={Journal of Scheduling},
  volume={19},
  number={4},
  pages={367--376},
  year={2016}
}

@ARTICLE{cho2022minimize,
  title={{Minimize Makespan of Permutation Flowshop Using Pointer Network}},
  author={Cho, Young In and Nam, So Hyun and Cho, Ki Young and Yoon, Hee Chang and Woo, Jong Hun},
  journal={Journal of Computational Design and Engineering},
  volume={9},
  number={1},
  pages={51--67},
  year={2022}
}

@ARTICLE{chung2025neural,
  title={{Neural Combinatorial Optimization with Reinforcement Learning in Industrial Engineering: A Survey}},
  author={Chung, K. T. and Lee, C. K. M. and Tsang, Y. P.},
  journal={Artificial Intelligence Review},
  volume={58},
  number={5},
  pages={130},
  year={2025}
}

@ARTICLE{davis2012smart,
  title={{Smart Manufacturing, Manufacturing Intelligence and Demand-Dynamic Performance}},
  author={Davis, Jim and Edgar, Thomas and Porter, James and Bernaden, John and Sarli, Michael},
  journal={Computers \& Chemical Engineering},
  volume={47},
  pages={145--156},
  year={2012}
}

@ARTICLE{fleszar2018algorithms,
  title={{Algorithms for the Unrelated Parallel Machine Scheduling Problem with a Resource Constraint}},
  author={Fleszar, Krzysztof and Hindi, Khalil S.},
  journal={European Journal of Operational Research},
  volume={271},
  number={3},
  pages={839--848},
  year={2018}
}

@ARTICLE{gedik2018constraint,
  title={{A Constraint Programming Approach for Solving Unrelated Parallel Machine Scheduling Problem}},
  author={Gedik, Ridvan and Kalathia, Darshan and Egilmez, Gokhan and Kirac, Emre},
  journal={Computers \& Industrial Engineering},
  volume={121},
  pages={139--149},
  year={2018}
}

@ARTICLE{kim2002unrelated,
  title={{Unrelated Parallel Machine Scheduling with Setup Times Using Simulated Annealing}},
  author={Kim, Dong-Won and Kim, Kyong-Hee and Jang, Wooseung and Chen, F. Frank},
  journal={Robotics and Computer-Integrated Manufacturing},
  volume={18},
  number={3-4},
  pages={223--231},
  year={2022}
}

@ARTICLE{lee1997heuristic,
  title={{A Heuristic to Minimize the Total Weighted Tardiness with Sequence-Dependent Setups}},
  author={Lee, Young Hoon and Bhaskaran, Kumar and Pinedo, Michael},
  journal={IISE Transactions},
  volume={29},
  number={1},
  pages={45--52},
  year={1997}
}

@ARTICLE{lee2013tabu,
  title={{A Tabu Search Algorithm for Unrelated Parallel Machine Scheduling with Sequence- and Machine-Dependent Setups: Minimizing Total Tardiness}},
  author={Lee, Jae-Ho and Yu, Jae-Min and Lee, Dong-Ho},
  journal={The International Journal of Advanced Manufacturing Technology},
  volume={69},
  pages={2081--2089},
  year={2013}
}

@ARTICLE{li2021hybrid,
  title={{A Hybrid Differential Evolution Algorithm for Parallel Machine Scheduling of Lace Dyeing Considering Colour Families, Sequence-Dependent Setup and Machine Eligibility}},
  author={Li, Debiao and Wang, Jing and Qiang, Rui and Chiong, Raymond},
  journal={International Journal of Production Research},
  volume={59},
  number={9},
  pages={2722--2738},
  year={2021}
}

@ARTICLE{li2022novel,
  title={{Novel Efficient Formulation and Matheuristic for Large-Sized Unrelated Parallel Machine Scheduling with Release Dates}},
  author={Li, Yantong and C{\^o}t{\'e}, Jean-Fran{\c{c}}ois and Coelho, Leandro C. and Wu, Peng},
  journal={International Journal of Production Research},
  volume={60},
  number={20},
  pages={6104--6123},
  year={2022}
}

@ARTICLE{lin2014unrelated,
  title={{Unrelated Parallel Machine Scheduling with Setup Times and Ready Times}},
  author={Lin, Yang-Kuei and Hsieh, Feng-Yu},
  journal={International Journal of Production Research},
  volume={52},
  number={4},
  pages={1200--1214},
  year={2014}
}

@ARTICLE{mazyavkina2021reinforcement,
  title={{Reinforcement Learning for Combinatorial Optimization: A Survey}},
  author={Mazyavkina, Nina and Sviridov, Sergey and Ivanov, Sergei and Burnaev, Evgeny},
  journal={Computers \& Operations Research},
  volume={134},
  pages={105400},
  year={2021}
}

@ARTICLE{mokotoff2001parallel,
  title={{Parallel Machine Scheduling Problems: A Survey}},
  author={Mokotoff, Ethel},
  journal={Asia-Pacific Journal of Operational Research},
  volume={18},
  number={2},
  pages={193--205},
  year={2001}
}

@ARTICLE{moser2022exact,
  title={{Exact and Metaheuristic Approaches for Unrelated Parallel Machine Scheduling}},
  author={Moser, Maximilian and Musliu, Nysret and Schaerf, Andrea and Winter, Felix},
  journal={Journal of Scheduling},
  volume={25},
  number={5},
  pages={507--534},
  year={2022}
}

@ARTICLE{mourtzis2022advances,
  title={{Advances in Adaptive Scheduling in Industry 4.0}},
  author={Mourtzis, Dimitris},
  journal={Frontiers in Manufacturing Technology},
  volume={2},
  pages={937889},
  year={2022}
}

@ARTICLE{park2021learning,
  title={{Learning to Schedule Job-Shop Problems: Representation and Policy Learning Using Graph Neural Network and Reinforcement Learning}},
  author={Park, Junyoung and Chun, Jaehyeong and Kim, Sang Hun and Kim, Youngkook and Park, Jinkyoo},
  journal={International Journal of Production Research},
  volume={59},
  number={11},
  pages={3360--3377},
  year={2021}
}

@ARTICLE{pei2020new,
  title={{A New Approximation Algorithm for Unrelated Parallel Machine Scheduling with Release Dates}},
  author={Pei, Zhi and Wan, Mingzhong and Wang, Ziteng},
  journal={Annals of Operations Research},
  volume={285},
  number={1},
  pages={397--425},
  year={2020}
}

@ARTICLE{pfund2004survey,
  title={{A Survey of Algorithms for Single and Multi-Objective Unrelated Parallel-Machine Deterministic Scheduling Problems}},
  author={Pfund, Michele and Fowler, John W. and Gupta, Jatinder N. D.},
  journal={Journal of the Chinese Institute of Industrial Engineers},
  volume={21},
  number={3},
  pages={230--241},
  year={2004}
}

@ARTICLE{rabadi2006heuristics,
  title={{Heuristics for the Unrelated Parallel Machine Scheduling Problem with Setup Times}},
  author={Rabadi, Ghaith and Moraga, Reinaldo J. and Al-Salem, Ameer},
  journal={Journal of Intelligent Manufacturing},
  volume={17},
  pages={85--97},
  year={2006}
}

@ARTICLE{raffin2021stable,
  title={{Stable-Baselines3: Reliable Reinforcement Learning Implementations}},
  author={Raffin, Antonin and Hill, Ashley and Gleave, Adam and Kanervisto, Anssi and Ernestus, Maximilian and Dormann, Noah},
  journal={Journal of Machine Learning Research},
  volume={22},
  number={268},
  pages={1--8},
  year={2021}
}

@ARTICLE{saracc2022mix,
  title={{A Mix Integer Programming Model and Solution Approach to Determine the Optimum Machine Number in the Unrelated Parallel Machine Scheduling Problem}},
  author={Sara{\c{c}}, Tu{\u{g}}ba and Tutumlu, Busra},
  journal={Journal of the Faculty of Engineering and Architecture of Gazi University},
  volume={37},
  number={1},
  pages={145--158},
  year={2022}
}

@ARTICLE{vallada2011genetic,
  title={{A Genetic Algorithm for the Unrelated Parallel Machine Scheduling Problem with Sequence Dependent Setup Times}},
  author={Vallada, Eva and Ruiz, Rub{\'e}n},
  journal={European Journal of Operational Research},
  volume={211},
  number={3},
  pages={612--622},
  year={2011}
}

@INPROCEEDINGS{cho2024reinforcement,
  title={{Reinforcement Learning for Unrelated Parallel Machine Scheduling with Release Dates, Setup Times, and Machine Eligibility}},
  author={Cho, Sang-Hyun and Kim, Hyun-Jung and M{\"o}nch, Lars},
  booktitle={{2024 Winter Simulation Conference (WSC)}},
  pages={1773--1784},
    year={2024}
}

@INPROCEEDINGS{cunha2020deep,
  title={{Deep Reinforcement Learning as a Job Shop Scheduling Solver: A Literature Review}},
  author={Cunha, Bruno and Madureira, Ana M. and Fonseca, Benjamim and Coelho, Duarte},
  booktitle={{Proceedings of the 18th International Conference on Hybrid Intelligent Systems (HIS 2018)}},
  pages={350--359},
  year={2020},
  publisher={Springer},
  address={Porto, Portugal},
  note={December 13-15, 2018}
}

@INPROCEEDINGS{huang2019review,
  title={{A Review of Combinatorial Optimization with Graph Neural Networks}},
  author={Huang, Tingfei and Ma, Yang and Zhou, Yuzhen and Huang, Honglan and Chen, Dongmei and Gong, Zidan and Liu, Yao},
  booktitle={{Proceedings of the 5th International Conference on Big Data and Information Analytics (BigDIA)}},
  pages={72--77},
  year={2019},
  note={December 20-22, Nanning, China}
}

@INPROCEEDINGS{kwon2021matrix,
  title={{Matrix Encoding Networks for Neural Combinatorial Optimization}},
  author={Kwon, Yeong-Dae and Choo, Jinho and Yoon, Iljoo and Park, Minah and Park, Duwon and Gwon, Youngjune},
  booktitle={{Advances in Neural Information Processing Systems}},
  volume={34},
  pages={5138--5149},
  year={2021}
}

@INPROCEEDINGS{nam2024,
  author={Nam, SoHyun and Baek, Jiwon and Cho, Young-In and Woo, Jong Hun},
  booktitle={{2024 Winter Simulation Conference (WSC)}}, 
  title={{Deep Reinforcement Learning for Setup and Tardiness Minimization in Parallel Machine Scheduling}}, 
  year={2024}
}

@INPROCEEDINGS{norman2024,
  author={Norman, David and Dawadi, Prafulla and Yedidsion, Harel},
  booktitle={{2024 Winter Simulation Conference (WSC)}}, 
  title={{Yield Improvement Using Deep Reinforcement Learning for Dispatch Rule Tuning}}, 
  year={2024}
}

@INPROCEEDINGS{soykanICMLA2023,
  author={Soykan, Bulent and Rabadi, Ghaith},
  booktitle={{Proceedings of the 2023 International Conference on Machine Learning and Applications (ICMLA)}}, 
  title={{Optimizing Multi Commodity Flow Problem Under Uncertainty: A Deep Reinforcement Learning Approach}}, 
  year={2023},
  pages={1267-1272},
  note={December 15-17, Jacksonville, Florida}
}

@INPROCEEDINGS{soykanWintersim2024,
  author={Soykan, Bulent and Rabadi, Ghaith},
  booktitle={{2024 Winter Simulation Conference (WSC)}}, 
  title={{Optimizing Job Shop Scheduling Problem Through Deep Reinforcement Learning and Discrete Event Simulation}}, 
  year={2024}
}

@INPROCEEDINGS{zhang2020learning,
  title={{Learning to Dispatch for Job Shop Scheduling via Deep Reinforcement Learning}},
  author={Zhang, Cong and Song, Wen and Cao, Zhiguang and Zhang, Jie and Tan, Puay Siew and Chi, Xu},
  booktitle={{Advances in Neural Information Processing Systems}},
  volume={33},
  pages={1621--1632},
  year={2020}
}

@MISC{brody2021attentive,
  title={{How Attentive Are Graph Attention Networks?}},
  author={Brody, Shaked and Alon, Uri and Yahav, Eran},
  journal={arXiv preprint arXiv:2105.14491},
  year={2021},
  howpublished = {\textit{arXiv preprint arXiv:2105.14491}}
}

@MISC{kool2018attention,
  title={{Attention, Learn to Solve Routing Problems!}},
  author={Kool, Wouter and Van Hoof, Herke and Welling, Max},
  journal={arXiv preprint arXiv:1803.08475},
  year={2018},
  howpublished = {\textit{arXiv preprint arXiv:1803.08475}}
}

@MISC{paszke2019pytorch,
  title={{PyTorch: An Imperative Style, High-Performance Deep Learning Library}},
  author={Paszke, Adam and Gross, Sam and Massa, Francisco and Lerer, Adam and Bradbury, James and Chanan, Gregory and Killeen, Trevor and Lin, Zeming and Gimelshein, Natalia and Antiga, Luca and others},
  booktitle={{Advances in Neural Information Processing Systems}},
  volume={32},
  year={2019}
}

@article{schulman2017proximal,
  title={Proximal policy optimization algorithms},
  author={Schulman, John and Wolski, Filip and Dhariwal, Prafulla and Radford, Alec and Klimov, Oleg},
  journal={arXiv preprint arXiv:1707.06347},
  year={2017}
}

\end{document}